%% file: main.tex
\documentclass[10pt,twocolumn,letterpaper]{article}

\usepackage[pagenumbers]{cvpr} 


\usepackage{hyperref}
\usepackage{amsmath}

\title{FlowMorph: Physics-Consistent Self-Supervision for Label-Free Single-Cell Mechanics in Microfluidic Videos}

\author{Bora Yimenicioglu\\
RareGen\\
Oakton, VA, USA
\and
Vishal Manikanden\\
Cornell University\\
Ithaca, NY, USA
}

\begin{document}
\maketitle
\input{sec/0_abstract}

{
    \small
    \bibliographystyle{ieeenat_fullname}
    \bibliography{main}
}


\end{document}

%% file: sec/0_abstract.tex
\begin{abstract}
Mechanical properties of red blood cells (RBCs) are promising biomarkers for hematologic and systemic disease, motivating microfluidic assays that probe deformability at throughputs of $10^3$--$10^6$ cells per experiment. However, existing pipelines rely on supervised segmentation or hand-crafted kymographs and rarely encode the laminar Stokes-flow physics that governs RBC shape evolution. We introduce \emph{FlowMorph}, a physics-consistent self-supervised framework that learns a label-free scalar mechanics proxy $k$ for each tracked RBC from short brightfield microfluidic videos. FlowMorph models each cell by a low-dimensional parametric contour, advances boundary points through a differentiable ``capsule-in-flow'' combining laminar advection and curvature-regularized elastic relaxation, and optimizes a loss coupling silhouette overlap, intra-cellular flow agreement, area conservation, wall constraints, and temporal smoothness, using only automatically derived silhouettes and optical flow.

Across four public RBC microfluidic datasets, FlowMorph achieves a mean silhouette IoU of $0.905$ on physics-rich videos with provided velocity fields and markedly improves area conservation and wall violations over purely data-driven baselines. On $\sim 1.5\times 10^5$ centered sequences, the scalar $k$ alone separates tank-treading from flipping dynamics with an AUC of $0.863$. Using only $200$ real-time deformability cytometry (RT-DC) events for calibration, a monotone map $E=g(k)$ predicts apparent Young's modulus with a mean absolute error of $0.118$\,MPa on $600$ held-out cells and degrades gracefully under shifts in channel geometry, optics, and frame rate.
\end{abstract}

\section{Introduction}
Mechanical phenotyping of single cells provides biophysical information that complements molecular markers and can reveal disease-related changes in membrane and cytoskeletal mechanics.\,\cite{otto2015_rtdc,chen2023_microfluidic_deformability}  In red blood cells (RBCs), deformability is tightly linked to disorders such as sickle cell disease, malaria, and inflammatory states, and transitions between tank-treading and flipping dynamics in shear flow are sensitive to these alterations.\,\cite{atwell2022_rbc_shear,darrin2023_redcell_dynamics}  Microfluidic deformability cytometry, including real-time deformability cytometry (RT-DC) and related constriction and shear-flow platforms, can probe $10^3$--$10^6$ cells per experiment and typically record brightfield videos from which deformation index and apparent Young's modulus are inferred.\,\cite{otto2015_rtdc,herbig2018_rtdc_stats}

Most analysis pipelines still treat these videos as generic image sequences.  Segmentation networks are trained on device-specific manual masks, and downstream mechanical descriptors are extracted from morphology alone (e.g., axis ratios, kymographs) without explicitly enforcing laminar Stokes-flow physics, area conservation, or no-slip at the walls.\,\cite{zhou2023_cv_meets_microfluidics,herbig2018_rtdc_stats,chen2023_microfluidic_deformability}  As a result, models can be brittle under changes in geometry, optical setup, or flow regime, and their latent representations are rarely interpretable as mechanical quantities.  At the other extreme, directly applying general-purpose physics-informed neural networks (PINNs) to full fluid–structure simulations is computationally demanding and difficult to integrate into high-throughput analysis.\,\cite{raissi2019_pinn}

\begin{figure}[t]
    \centering
    \includegraphics[width=\linewidth]{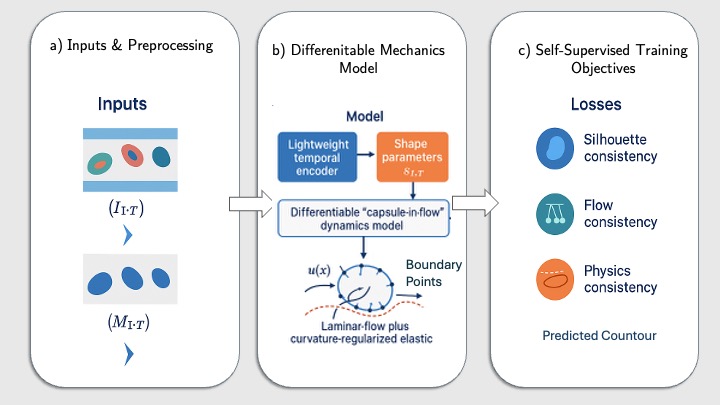}
    \caption{\textbf{FlowMorph overview.} Short brightfield clips of red blood cells (RBCs) flowing through microfluidic channels are converted into coarse silhouettes and optical flow.  A lightweight encoder and temporal smoother infer a low-dimensional parametric contour and a per-track scalar mechanics proxy $k$.  A differentiable physics layer advects boundary points under laminar flow plus curvature-regularized elastic relaxation, rasterizes the predicted contour into a soft mask, and computes self-supervised losses enforcing silhouette consistency, intra-cellular flow agreement, area conservation, wall constraints, and temporal smoothness.  After training, a monotone map $E=g(k)$ calibrated on a small subset of real-time deformability cytometry (RT-DC) events yields approximate Young's modulus estimates.}
    \label{fig:pipeline}
\end{figure}

We address these gaps with \emph{FlowMorph}, a physics-consistent self-supervised framework that learns a label-free scalar mechanics proxy $k$ for each tracked RBC from short brightfield clips (Fig.~\ref{fig:pipeline}).  FlowMorph infers a low-dimensional parametric contour, advances boundary points through a differentiable ``capsule-in-flow'' model that combines measured or locally modeled laminar velocity fields with curvature-regularized elastic relaxation, and rasterizes the predicted contour into a soft mask.  The training objective couples silhouette overlap, intra-cellular optical-flow agreement, area conservation, wall constraints when channel masks and velocity fields are available, and temporal smoothness of both contour parameters and the per-track scalar $k$, using only automatically derived silhouettes and optical flow.\,\cite{cimrak2023_rbc_tracking}  A separate tiny calibration step maps $k$ to an apparent Young's modulus $E$ using a small subset of RT-DC measurements with hyperelastic lookup tables.\,\cite{otto2015_rtdc,herbig2018_rtdc_stats}  Experiments on four public datasets show that FlowMorph improves physics-validity metrics, yields a strong mechanics signal for tank-treading versus flipping dynamics, and, after tiny-set calibration, produces approximate Young's modulus estimates that remain robust under domain shifts in geometry, optics, and frame rate.\,\cite{cimrak2023_rbc_tracking,darrin2023_redcell_dynamics,bentley2022_confinement}

\section{Related Work}
Microfluidic deformability cytometry platforms such as real-time deformability cytometry (RT-DC) and related constriction and extensional-flow devices enable label-free mechanical phenotyping of $10^3$--$10^6$ cells per experiment by imaging cells in microchannels and deriving descriptors such as deformation index, relaxation time, and apparent Young's modulus.\,\cite{otto2015_rtdc,chen2023_microfluidic_deformability,herbig2018_rtdc_stats}  In red blood cells (RBCs), these assays probe shear-induced transitions between tank-treading and flipping and are sensitive to membrane and cytoskeletal alterations caused by disease.\,\cite{atwell2022_rbc_shear,darrin2023_redcell_dynamics}  Most existing pipelines, however, extract mechanics from hand-crafted kymographs or axis ratios and rely on supervised segmentation tuned to specific devices and imaging conditions, limiting robustness across labs and geometries.\,\cite{otto2015_rtdc,chen2023_microfluidic_deformability,zhou2023_cv_meets_microfluidics}

Curated microfluidic datasets have catalyzed the use of modern computer vision methods in single-cell mechanics.  Cimr{\'a}k \emph{et al.} provide brightfield RBC videos with per-pixel velocity fields for benchmarking laminar-flow tracking algorithms.\,\cite{cimrak2023_rbc_tracking}  Darrin \emph{et al.} assemble $\sim 1.5\times10^5$ centered RBC sequences with tank-treading and flipping labels and show that convolutional and recurrent architectures can classify dynamic behaviors relevant to sickle cell disease.\,\cite{darrin2023_redcell_dynamics}  Bentley \emph{et al.} study single-cell motility in confinement and report behavioral heterogeneity arising from hydrodynamics and geometry.\,\cite{bentley2022_confinement}  Zhou \emph{et al.} review label-free microfluidic cell analysis with classical image processing and supervised deep networks, highlighting sensitivity to domain shift and the absence of explicit fluid-mechanics constraints.\,\cite{zhou2023_cv_meets_microfluidics,zhou2024_ai_microfluidics}  These works primarily treat microfluidic videos as generic image sequences: deep networks are trained end-to-end on pixels and labels, without embedding laminar-flow or capsule-mechanics constraints.

Self-supervised methods for live-cell microscopy exploit temporal coherence and motion cues to train segmentation models without human masks, typically enforcing consistency between predictions and frame-to-frame optical flow or synthetic perturbations.\,\cite{robitaille2022_selfsup_segmentation}  In optical flow, unsupervised approaches such as UnFlow leverage forward–backward photometric consistency and robust losses to learn dense fields without ground truth,\,\cite{meister2018_unflow} while physics-informed neural networks (PINNs) and differentiable simulators show that embedding PDE residuals or differentiable dynamics into the learning loop can improve data efficiency and extrapolation in fluid mechanics and elasticity.\,\cite{raissi2019_pinn,sanchez2020_graphsim}  FlowMorph differs in that (i) it learns a per-cell scalar mechanics proxy $k$ without any human masks, motion labels, or stiffness measurements, using self-supervised losses defined on automatically derived silhouettes, optical flow, and channel masks where available;\,\cite{cimrak2023_rbc_tracking,robitaille2022_selfsup_segmentation} (ii) it employs a minimal differentiable ``capsule-in-flow'' model that explicitly encodes laminar velocity fields and curvature-regularized elastic relaxation at the level of individual cell contours rather than full-field simulations;\,\cite{otto2015_rtdc,atwell2022_rbc_shear,raissi2019_pinn} and (iii) it is calibrated to physical stiffness via a tiny subset of RT-DC events and evaluated for cross-domain robustness across channel geometries, optical setups, and frame rates.\,\cite{herbig2018_rtdc_stats,bentley2022_confinement}

\section{Method}
\label{sec:method}
FlowMorph combines a low-dimensional contour representation, a differentiable ``capsule-in-flow'' dynamics model, and a lightweight temporal encoder to infer a per-track scalar mechanics proxy $k$ from short brightfield clips.  Given a tracked sequence $(I_{1:T})$ and coarse silhouettes $(M_{1:T})$ extracted automatically, the model predicts shape parameters $s_{1:T}$ and a scalar $k$, advances boundary points under laminar flow plus curvature-regularized elastic relaxation, rasterizes the predicted contour into a soft mask, and optimizes self-supervised losses that enforce silhouette, flow, and physics consistency.

\subsection{Overview and Notation}
We consider short brightfield sequences in which individual red blood cells (RBCs) flow through straight or weakly constricted microchannels under low-Reynolds-number conditions, so that the surrounding plasma can be modeled as an incompressible Newtonian fluid in the Stokes regime.\,\cite{otto2015_rtdc,chen2023_microfluidic_deformability,atwell2022_rbc_shear}  For each automatically tracked RBC we extract a cropped sequence
\(
I_{1:T} = (I_1,\dots,I_T)
\)
and derive coarse binary silhouettes $M_t$ by thresholding and morphological cleanup; these masks are treated as noisy observations for self-supervision rather than ground truth.\,\cite{robitaille2022_selfsup_segmentation,cimrak2023_rbc_tracking}

When available (RBCdataset), we also have a binary channel mask $\Omega_{\text{wall}} \subset \mathbb{R}^2$ and a per-pixel velocity field $u(\mathbf{x}) \in \mathbb{R}^2$ obtained from Stokes simulations calibrated to the microchannel geometry.\,\cite{cimrak2023_rbc_tracking,chen2023_microfluidic_deformability}  In datasets where only centered crops are provided (RBC dynamics), we approximate the local flow by a simple extensional velocity gradient with learnable strain rate (Section~\ref{subsec:capsule_dynamics}).\,\cite{darrin2023_redcell_dynamics}

At each time $t$, the RBC outline is represented by a compact parameter vector $s_t \in \mathbb{R}^d$ and a set of boundary points $\{\mathbf{p}_t^n\}_{n=1}^N \subset \mathbb{R}^2$ sampled at fixed angles.  Each track is associated with a single scalar $k \in \mathbb{R}_{+}$, interpreted as an effective stiffness proxy and assumed to be constant over the $T$ frames of that track, reflecting the fact that RBC mechanical properties do not vary appreciably on these short time scales.\,\cite{atwell2022_rbc_shear,darrin2023_redcell_dynamics}

Our formulation relies on standard assumptions for RBCs in microfluidic channels: (i) Reynolds numbers $\mathrm{Re}\ll 1$ so inertia is negligible; (ii) flow profiles are near-Poiseuille in straight channels and well approximated locally by extensional shear in constrictions; (iii) RBCs are effectively incompressible, so projected area changes are small absent strong out-of-plane motion; and (iv) no-slip and non-penetration conditions hold at channel walls when present.\,\cite{otto2015_rtdc,chen2023_microfluidic_deformability,atwell2022_rbc_shear}  These assumptions are satisfied by the devices and flow conditions in the datasets we study.\,\cite{cimrak2023_rbc_tracking,darrin2023_redcell_dynamics,herbig2018_rtdc_stats,bentley2022_confinement}

\subsection{Shape Parameterization}
We represent RBC contours using a star-convex Fourier parameterization that captures moderate deviations from ellipticity while remaining low-dimensional and differentiable.\,\cite{atwell2022_rbc_shear,darrin2023_redcell_dynamics}  Let $(x_t,y_t)$ denote the centroid and $\phi\in[0,2\pi)$ the polar angle.  The radial distance from centroid to boundary is modeled as
\begin{equation}
    r_t(\phi) = c_{0,t} + \sum_{m=1}^{M} c_{m,t} \cos(m\phi) + d_{m,t} \sin(m\phi),
\end{equation}
with $M\leq 6$ and parameters
\(
s_t = [x_t,y_t,\{c_{m,t},d_{m,t}\}_{m=0}^{M}] .
\)
Boundary points are sampled at fixed angles $\phi_n = 2\pi n/N$ via
\begin{equation}
    \mathbf{p}_t^n =
    \begin{bmatrix}
    x_t \\ y_t
    \end{bmatrix}
    +
    r_t(\phi_n)
    \begin{bmatrix}
    \cos\phi_n \\ \sin\phi_n
    \end{bmatrix},
    \quad n=1,\dots,N.
\end{equation}
This yields a differentiable mapping $s_t \mapsto \{\mathbf{p}_t^n\}$ and admits finite-difference curvature estimates along the discretized contour.  In ablations we also consider a simpler ellipse model, but the star-convex parameterization is used for all main results.

\subsection{Capsule-in-Flow Dynamics}
\label{subsec:capsule_dynamics}
Given boundary points $\{\mathbf{p}_t^n\}$ at time $t$, we model their evolution over one frame interval $\Delta t$ as
\begin{equation}
    \mathbf{p}_{t+1}^n = \mathbf{p}_t^n + \Delta t \left( u(\mathbf{p}_t^n) + f_{\text{elastic}}(\mathbf{p}_t^n; k) + f_{\text{wall}}(\mathbf{p}_t^n) \right),
    \label{eq:boundary_update}
\end{equation}
where $u(\cdot)$ is the laminar velocity field, $f_{\text{elastic}}$ is an effective elastic force derived from a curvature-based energy with stiffness $k$, and $f_{\text{wall}}$ enforces non-penetration and approximate no-slip at the walls when $\Omega_{\text{wall}}$ is available.\,\cite{otto2015_rtdc,chen2023_microfluidic_deformability}

\paragraph{Advection by laminar flow.}
When a per-pixel velocity field $u(\mathbf{x})$ is provided, we treat it as given and sample $u(\mathbf{p}_t^n)$ directly.\,\cite{cimrak2023_rbc_tracking,chen2023_microfluidic_deformability}  When only centered crops are available (RBC dynamics), we approximate $u$ locally by a linear extensional flow,
\begin{equation}
u(\mathbf{x}) \approx 
\begin{bmatrix}
\epsilon & 0 \\
0 & -\epsilon
\end{bmatrix}
(\mathbf{x} - \mathbf{x}_\mathrm{c}) + \mathbf{b},
\label{eq:extensional_flow_method}
\end{equation}
where $\mathbf{x}_\mathrm{c}$ is the cell centroid, $\epsilon$ is a learnable strain rate constrained to a plausible range based on experimentally reported shear rates, and $\mathbf{b}$ is a translation term capturing bulk advection.\,\cite{darrin2023_redcell_dynamics,chen2023_microfluidic_deformability}  We regularize $\epsilon$ using a penalty that discourages strain rates inconsistent with the observed center-of-mass motion.

\paragraph{Curvature-regularized elastic relaxation.}
We model the RBC membrane as a thin shell with an effective bending energy that penalizes curvature fluctuations along the contour,\cite{atwell2022_rbc_shear,otto2015_rtdc}
\begin{equation}
    E_{\text{curve}}(s_t; k) = \frac{k}{2} \int_{\Gamma_t} \kappa(s)^2 \, ds,
    \label{eq:curve_energy}
\end{equation}
where $\Gamma_t$ is the contour at time $t$ and $\kappa$ is scalar curvature.  In discrete form, we approximate curvature at point $n$ via central finite differences on the sequence $\{\mathbf{p}_t^{n-1}, \mathbf{p}_t^n, \mathbf{p}_t^{n+1}\}$ and compute a discrete energy $E_{\text{curve}}(s_t; k)\approx \frac{k}{2}\sum_n \kappa_t(n)^2\Delta s$ with uniform spacing $\Delta s$.  The elastic force is then
\begin{equation}
    f_{\text{elastic}}(\mathbf{p}_t^n; k) = - \eta \, \nabla_{\mathbf{p}_t^n} E_{\text{curve}}(s_t; k),
\end{equation}
with step-size parameter $\eta>0$.  We implement this gradient via automatic differentiation on the discrete energy, ensuring that the mapping from $(s_t,k)$ to $\{\mathbf{p}_{t+1}^n\}$ remains differentiable.\,\cite{raissi2019_pinn}  Intuitively, larger $k$ produces stronger curvature-regularized relaxation toward smoother contours, so $k$ serves as an effective stiffness proxy opposing flow-induced deformation.

\paragraph{Wall non-penetration and no-slip.}
When a channel mask $\Omega_{\text{wall}}$ is available, we impose non-penetration and approximate no-slip at the walls.\,\cite{otto2015_rtdc,chen2023_microfluidic_deformability}  We precompute a signed distance field $\phi_{\text{wall}}(\mathbf{x})$ that is negative inside the walls and positive in the fluid and add a soft potential that penalizes boundary points with $\phi_{\text{wall}}(\mathbf{p}_t^n)<0$.  A small additional penalty discourages tangential motion along the wall within a narrow band around $\partial\Omega_{\text{wall}}$, implementing an approximate no-slip condition.  These contributions are summarized in a wall loss term $\mathcal{L}_{\text{noslip}}$ in Section~\ref{subsec:losses}.  A schematic of the one-step update and loss paths is provided in the supplement (Fig.~S1).

\subsection{Differentiable Rasterization}
To compare the predicted contour to the coarse silhouette $M_{t+1}$ we convert boundary points to a soft segmentation mask $\hat{M}_{t+1}$.  We construct a signed distance field $d(\mathbf{x}; s_{t+1})$ on the cropped region by computing, for each pixel center $\mathbf{x}$, the signed distance to the polygonal contour defined by $\{\mathbf{p}_{t+1}^n\}$.  Distances are mapped to occupancy probabilities via
\begin{equation}
    \hat{M}_{t+1}(\mathbf{x}) = \sigma\left(-\frac{d(\mathbf{x}; s_{t+1})}{\tau}\right),
    \label{eq:soft_mask}
\end{equation}
where $\sigma$ is the sigmoid and $\tau>0$ is a temperature that controls boundary softness.  Small $\tau$ yields near-binary masks; larger $\tau$ stabilizes gradients near the contour.  The mapping $s_{t+1} \mapsto \hat{M}_{t+1}$ is differentiable almost everywhere, allowing gradients of the self-supervised losses to flow back to both the contour parameters and $k$.

\subsection{Temporal Inference of $s_{1:T}$ and $k$}
Given a tracked clip $(I_{1:T})$ and optional channel mask $\Omega_{\text{wall}}$, our goal is to infer both the contour sequence $s_{1:T}$ and the per-track scalar $k$.  Each frame $I_t$ is passed through a small convolutional encoder $E_{\theta}$ (three convolutional blocks with ReLU) to produce a feature vector $z_t = E_{\theta}(I_t) \in \mathbb{R}^{h}$, optionally concatenated with coarse geometric features derived from $M_t$ (area and centroid).\,\cite{darrin2023_redcell_dynamics}

The feature sequence $(z_{1:T})$ is processed by a gated recurrent unit (GRU)\,\cite{cho2014_gru}
\begin{equation}
    h_t = \mathrm{GRU}(z_t, h_{t-1}), \quad t = 1,\dots,T,
\end{equation}
with hidden dimension 64--128.  From each $h_t$ we predict a residual update $\Delta s_t$ to the contour parameters and an intermediate stiffness proposal $\tilde{k}_t$.  Contour parameters are updated via $s_t = s_{t-1} + \Delta s_t$, initialized by an ellipse fit to $M_1$.  A track-wise scalar is obtained by averaging proposals,
\begin{equation}
    k = \frac{1}{T} \sum_{t=1}^{T} \tilde{k}_t,
\end{equation}
followed by a softplus nonlinearity to ensure positivity.  A temporal smoothness term in the loss encourages $\tilde{k}_t$ to be nearly constant over the track, consistent with the assumption of track-wise constant mechanics.\,\cite{atwell2022_rbc_shear,darrin2023_redcell_dynamics}  In ablations we replace the GRU with a linear-Gaussian state-space model on $s_t$ and treat $k$ as a global latent variable optimized directly.

\subsection{Self-Supervised Objective}
\label{subsec:losses}
The model is trained end-to-end to minimize a composite loss that enforces consistency between predicted and observed silhouettes, intra-cellular motion, area, wall constraints, and temporal smoothness, without using any human-provided labels or mechanics measurements.

\paragraph{Silhouette consistency.}
We measure agreement between the predicted soft mask $\hat{M}_{t+1}$ and the coarse silhouette $M_{t+1}$ via a soft intersection-over-union (IoU),
\begin{equation}
\begin{split}
\mathrm{IoU}(\hat{M}_{t+1}, M_{t+1}) =
\frac{\sum_{\mathbf{x}} \hat{M}_{t+1}(\mathbf{x}) M_{t+1}(\mathbf{x})}
{\sum_{\mathbf{x}} \big(
\hat{M}_{t+1}(\mathbf{x}) + M_{t+1}(\mathbf{x})
- \hat{M}_{t+1}(\mathbf{x}) M_{t+1}(\mathbf{x})
\big)}.
\end{split}
\end{equation}

\paragraph{Optical-flow agreement.}
Let $v_t^{\text{obs}}(\mathbf{x})$ denote the observed optical flow from $I_t$ to $I_{t+1}$, estimated on the cropped region.\,\cite{meister2018_unflow}  The model predicts a per-pixel velocity inside the cell by combining the laminar field and elastic forces; we compare them via
\begin{equation}
    \mathcal{L}_{\text{flow}} =
    \sum_{\mathbf{x}} \hat{M}_t(\mathbf{x}) \,
    \left\| v_t^{\text{obs}}(\mathbf{x}) - v_t^{\text{pred}}(\mathbf{x}) \right\|_1.
\end{equation}

\paragraph{Area conservation.}
Assuming incompressibility, the projected area of an RBC should remain approximately constant over the time scale of a track.\,\cite{atwell2022_rbc_shear,otto2015_rtdc}  Defining $A_t = \sum_{\mathbf{x}} \hat{M}_t(\mathbf{x})$, we penalize area deviations via
\begin{equation}
    \mathcal{L}_{\text{area}} = \sum_{t=1}^{T-1} (A_{t+1} - A_t)^2.
\end{equation}

\paragraph{Wall constraints and temporal smoothness.}
When a channel mask is available, the non-penetration and no-slip penalties described above contribute a wall loss $\mathcal{L}_{\text{noslip}}$ that discourages boundary points inside the walls and tangential motion along $\partial\Omega_{\text{wall}}$.\,\cite{otto2015_rtdc,chen2023_microfluidic_deformability}  To regularize dynamics, we add
\begin{equation}
    \mathcal{L}_{\text{smooth}} = \sum_{t=2}^{T} \|s_t - s_{t-1}\|_2^2 + \alpha \sum_{t=2}^{T} \|\tilde{k}_t - \tilde{k}_{t-1}\|_2^2,
\end{equation}
with small $\alpha>0$, encouraging smooth evolution of $s_t$ and discouraging frame-to-frame fluctuations in $\tilde{k}_t$ while allowing different tracks to have distinct stiffness proxies.\,\cite{atwell2022_rbc_shear,darrin2023_redcell_dynamics}

\paragraph{Photometric consistency and total loss.}
On sequences with sufficient contrast, we add a photometric loss inside the predicted mask by warping $I_t$ with $v_t^{\text{pred}}$ and comparing to $I_{t+1}$ using a Charbonnier penalty
\begin{equation}
    \mathcal{L}_{\text{photo}} = \sum_{\mathbf{x}} \hat{M}_t(\mathbf{x}) \,
    \rho\left( I_{t+1}(\mathbf{x}) - I_{t}( \mathbf{x} + v_t^{\text{pred}}(\mathbf{x}) ) \right),
\end{equation}
where $\rho(r) = \sqrt{r^2 + \epsilon^2}$ with small $\epsilon>0$.  
The full self-supervised loss for a track is
\begin{equation}
\begin{split}
    \mathcal{L} =
    \sum_{t=1}^{T-1} \Big(
      w_1 \mathcal{L}_{\text{IoU}}
    + w_2 \mathcal{L}_{\text{flow}}
    + w_3 \mathcal{L}_{\text{area}} \\
    + w_4 \mathcal{L}_{\text{noslip}}
    + w_5 \mathcal{L}_{\text{smooth}}
    + w_6 \mathcal{L}_{\text{photo}}
    \Big).
\end{split}
\end{equation}

with weights $(w_1,\dots,w_6)$ set to $(1.0, 0.5, 0.2, 0.5, 0.05, 0.1)$ in our main experiments and tuned in a small grid search on validation sequences.  On datasets without channel masks or reliable photometric information, the corresponding terms are disabled.  Optimizing this physics-consistent loss yields temporally coherent contours and a per-track scalar $k$ that capture the strength of curvature-regularized relaxation consistent with observed motion, without any manual masks, motion labels, or mechanical measurements during training.

\section{Datasets and Evaluation Protocol}
\label{sec:datasets}
We evaluate FlowMorph on four public resources that are widely used in microfluidics and RBC mechanics: (i) microchannel videos with companion velocity fields, (ii) a large corpus of centered RBC dynamics sequences with motion labels, (iii) real-time deformability cytometry (RT-DC) measurements with apparent Young's modulus values from a hyperelastic lookup table, and (iv) a motility-in-confinement dataset.\,\cite{otto2015_rtdc,chen2023_microfluidic_deformability,cimrak2023_rbc_tracking,darrin2023_redcell_dynamics,bentley2022_confinement,herbig2018_rtdc_stats}  All data consist of de-identified in vitro recordings without patient-identifying information, and ethical approvals were obtained by the original authors; we restrict our work to secondary analysis, so no additional institutional review board (IRB) review is required.\,\cite{cimrak2023_rbc_tracking,darrin2023_redcell_dynamics,bentley2022_confinement,herbig2018_rtdc_stats}

\subsection{Datasets}
\paragraph{RBC tracking dataset (RBCdataset).}
RBCdataset contains brightfield videos of RBCs flowing through an $80~\mu$m U-channel and a stenosed channel, along with per-pixel velocity fields obtained from Stokes simulations calibrated to the device geometry.\,\cite{cimrak2023_rbc_tracking}  We use this dataset to train and evaluate all physics-based loss terms involving the flow field and wall constraints.  After flat-field correction and background subtraction, we detect and track RBCs, extract short ROIs around each track, derive coarse silhouettes by thresholding and morphology, and estimate optical flow using an UnFlow-style multi-scale method.\,\cite{meister2018_unflow}  Tracks are split into training, validation, and test sets at the video level to avoid leakage.

\paragraph{RBC dynamics dataset.}
The Scientific Reports dataset of Darrin \emph{et al.} contains $\sim 1.5\times10^5$ centered RBC sequences acquired under controlled shear flow, each labeled as tank-treading, flipping, or unreliable based on dynamic behavior.\,\cite{darrin2023_redcell_dynamics,atwell2022_rbc_shear}  Each sequence is a $31\times31$ crop of varying length centered on a single RBC.  We treat these as local laminar-flow patches without explicit wall geometry and approximate the surrounding flow by the extensional model in Eq.~\eqref{eq:extensional_flow_method}.  Coarse silhouettes and optical flow are computed per sequence, and we use only TT and FL sequences, splitting experiments into training, validation, and test sets to probe cross-condition generalization.

\paragraph{RT-DC subset for calibration.}
RT-DC measures cell deformation during transit through a constriction and uses a hyperelastic lookup table to map deformation and size to an apparent Young's modulus $E$.\,\cite{otto2015_rtdc,herbig2018_rtdc_stats}  From the RT-DC statistics dataset, we sample 800 RBC events with valid deformation and $E$ values, reserving 200 cells for calibration and 600 for held-out testing.\,\cite{herbig2018_rtdc_stats}  After training FlowMorph, we treat each image as a single-frame track, infer a scalar $k$, and fit a monotone mapping $E=g(k)$ by isotonic regression on the calibration set.\,\cite{barlow1972_isotonic}  This mapping is used only at test time and is never backpropagated through the main model.

\paragraph{Confinement motility dataset.}
Bentley \emph{et al.} study single-cell motility in microfluidic confinement using maze-like channels that generate complex trajectories governed by hydrodynamics and geometry.\,\cite{bentley2022_confinement}  We use their videos and tracking code to extract RBC tracks and apply the same preprocessing pipeline as for RBCdataset, but without velocity fields or channel masks.  This dataset is used exclusively as an out-of-distribution (OOD) test for the learned mechanics proxy $k$ and physics-validity metrics.

\subsection{Evaluation Protocol and Metrics}
We evaluate FlowMorph along four axes: (i) physics validity and segmentation quality on physics-rich videos; (ii) mechanics signal for discriminating dynamic behaviors; (iii) calibration to physical stiffness; and (iv) robustness to domain shifts and training stochasticity.  Unless otherwise noted, all numbers are averaged over five independent training runs with different random seeds, and 95\% confidence intervals (CIs) are estimated by nonparametric bootstrap.\,\cite{herbig2018_rtdc_stats,efron1994_bootstrap}

On RBCdataset, we report mean soft IoU between $\hat{M}_t$ and $M_t$, the fraction of steps with relative projected area change below 2\%, the wall-violation rate (percentage of time steps with any boundary point inside the wall region), and the mean endpoint error (EPE) between observed and predicted intra-cellular flow inside the predicted mask.\,\cite{cimrak2023_rbc_tracking,chen2023_microfluidic_deformability,meister2018_unflow}  On the RBC dynamics test experiments, we treat $k$ as a continuous score for dynamic behavior and measure AUC for tank-treading vs.\ flipping, Spearman rank correlation between $k$ and the binary label, and expected calibration error (ECE) computed over deciles of predicted scores.\,\cite{darrin2023_redcell_dynamics,atwell2022_rbc_shear,herbig2018_rtdc_stats}  On the RT-DC subset, we evaluate mean absolute error (MAE) in apparent Young's modulus, negative log-likelihood (NLL) under a Gaussian noise model with variance estimated from calibration residuals, ECE over bins of predicted $E$, and the percentage of monotonicity violations, i.e., pairs $(i,j)$ with $k_i<k_j$ but $\hat{E}_i>\hat{E}_j$.\,\cite{otto2015_rtdc,herbig2018_rtdc_stats,barlow1972_isotonic}

To probe domain shift, we consider scenarios where we train on one microchannel geometry and test on another within RBCdataset, train on one optical setup and test on another within the RBC dynamics experiments, and evaluate physics-validity metrics and qualitative behavior of $k$ on the confinement dataset.\,\cite{cimrak2023_rbc_tracking,darrin2023_redcell_dynamics,bentley2022_confinement}  We report changes in AUC, MAE$(E)$, area-conservation rate, and wall-violation rate relative to in-domain performance (see Supplementary Table~S1). Robustness to training stochasticity is quantified by the coefficient of variation of the mean $k$ across runs.  Baselines include a classical deformation-index (DI) pipeline, a mask-heavy deep model that combines Cellpose or Mask R-CNN masks with a supervised regressor, and a no-physics network that shares FlowMorph's encoder and temporal backbone but predicts next-frame masks via a learned warp without explicit laminar-flow or elastic-relaxation terms; ablations remove individual loss terms or alter the contour and temporal models, and are evaluated with the same metrics.\,\cite{otto2015_rtdc,zhou2023_cv_meets_microfluidics,robitaille2022_selfsup_segmentation,stringer2021_cellpose,he2017_maskrcnn}

\section{Experiments}
\label{sec:experiments}

\subsection{Implementation Details}
We implement FlowMorph in PyTorch with automatic differentiation for all contour and physics operations.\,\cite{raissi2019_pinn}  For each tracked RBC, we extract a $64\times64$ region of interest (ROI) centered on the cell, normalize intensities per sequence, derive coarse silhouettes $M_t$ by thresholding and morphology, and estimate optical flow $v_t^{\text{obs}}$ using a multi-scale UnFlow-style method.\,\cite{cimrak2023_rbc_tracking,darrin2023_redcell_dynamics,meister2018_unflow}  Unless otherwise noted, we use a star-convex contour with $M=4$ Fourier modes and $N=128$ boundary points, a GRU with 128 hidden units, and three convolutional blocks in the spatial encoder.\,\cite{atwell2022_rbc_shear,cho2014_gru}. A detailed compute and throughput summary is provided in Supplementary Table~S3.

Training proceeds in three phases: a synthetic warm-start for the physics module and rasterizer, physics-rich training on RBCdataset with all loss terms active, and fine-tuning on the RBC dynamics dataset with the extensional flow model and wall losses disabled.\,\cite{otto2015_rtdc,chen2023_microfluidic_deformability,cimrak2023_rbc_tracking,darrin2023_redcell_dynamics}  RT-DC calibration is performed afterward and does not backpropagate through the main model.\,\cite{otto2015_rtdc,herbig2018_rtdc_stats,barlow1972_isotonic}  We optimize with Adam (learning rate $2\times10^{-4}$, cosine decay), batch size 16 tracks $\times$ 8 frames, gradient clipping at norm 1.0, and early stopping on validation physics metrics.\,\cite{darrin2023_redcell_dynamics}  On a single RTX~4090 GPU (24\,GB), synthetic warm-start, RBCdataset training, and RBC dynamics fine-tuning together require $\sim$17.2\,GPU-hours; at inference, FlowMorph processes $\approx$220 tracks/s on RBCdataset and $\approx$300 tracks/s on centered $31\times31$ sequences.  All results are averaged over five runs with different seeds, and 95\% confidence intervals are estimated by nonparametric bootstrap.\,\cite{herbig2018_rtdc_stats,efron1994_bootstrap}. To facilitate reproducibility, if this work is accepted we will release the FlowMorph source code, including preprocessing scripts, training and evaluation pipelines, configuration files, and isotonic calibration routines.

\subsection{Physics-Rich Microfluidic Videos}
We first evaluate FlowMorph on RBCdataset, where per-pixel laminar velocity fields and channel masks are available and all physics-based losses are active.\,\cite{cimrak2023_rbc_tracking}  Table~\ref{tab:physics_rbc} compares FlowMorph to a classical deformation-index (DI) pipeline, a mask-heavy deep model (Mask R-CNN + warp), and a no-physics network with the same encoder and GRU but a purely learned warp.\,\cite{otto2015_rtdc,chen2023_microfluidic_deformability,zhou2023_cv_meets_microfluidics,he2017_maskrcnn}  FlowMorph attains the highest silhouette IoU and area-conservation rate and the lowest wall-violation rate and intra-cellular flow error; for example, it reduces wall violations by nearly an order of magnitude relative to the no-physics network while also improving agreement with observed intra-cellular flow.  Mask R-CNN improves segmentation but still incurs more wall violations and higher flow error, suggesting that accurate mechanics cannot be recovered by segmentation alone.\,\cite{zhou2023_cv_meets_microfluidics}

The left panel of Fig.~\ref{fig:results} illustrates representative predictions on RBCdataset: FlowMorph follows laminar streamlines in both U-channel and constriction geometries while preserving projected area and avoiding wall penetration, whereas the no-physics baseline frequently produces contours that intersect the walls or exhibit unphysical shape changes.

\begin{table*}[t]
    \centering
    \caption{\textbf{Physics-rich RBCdataset results.}  Mean $\pm$95\% CI over 5 runs.  FlowMorph improves silhouette IoU, area conservation, wall non-penetration, and agreement with observed intra-cellular flow compared to purely data-driven baselines.}
    \label{tab:physics_rbc}
    \begin{tabular}{lcccc}
        \toprule
        Method & IoU $\uparrow$ & Area $\Delta{<}2\%\uparrow$ & Wall viol.\ $\downarrow$ & Flow EPE $\downarrow$ \\
        \midrule
        FlowMorph (ours) & $\mathbf{0.905}\pm0.006$ & $\mathbf{95.7}\pm1.2\,\%$ & $\mathbf{0.32}\pm0.09\,\%$ & $\mathbf{0.28}\pm0.03$ \\
        No-Physics Net & $0.861\pm0.009$ & $86.5\pm1.8\,\%$ & $2.94\pm0.31\,\%$ & $0.46\pm0.05$ \\
        Mask R-CNN + warp & $0.874\pm0.007$ & $89.1\pm1.5\,\%$ & $1.87\pm0.24\,\%$ & $0.39\pm0.04$ \\
        Optical-flow-only warp & $0.842\pm0.010$ & $78.3\pm2.1\,\%$ & $4.08\pm0.37\,\%$ & $0.64\pm0.07$ \\
        Classical DI pipeline & $0.803\pm0.011$ & $73.9\pm2.5\,\%$ & $5.22\pm0.42\,\%$ & $0.79\pm0.08$ \\
        \bottomrule
    \end{tabular}
\end{table*}

\subsection{Mechanics Signal and Dynamic Behavior}
\begin{figure}[t]
    \centering
    \includegraphics[width=\linewidth]{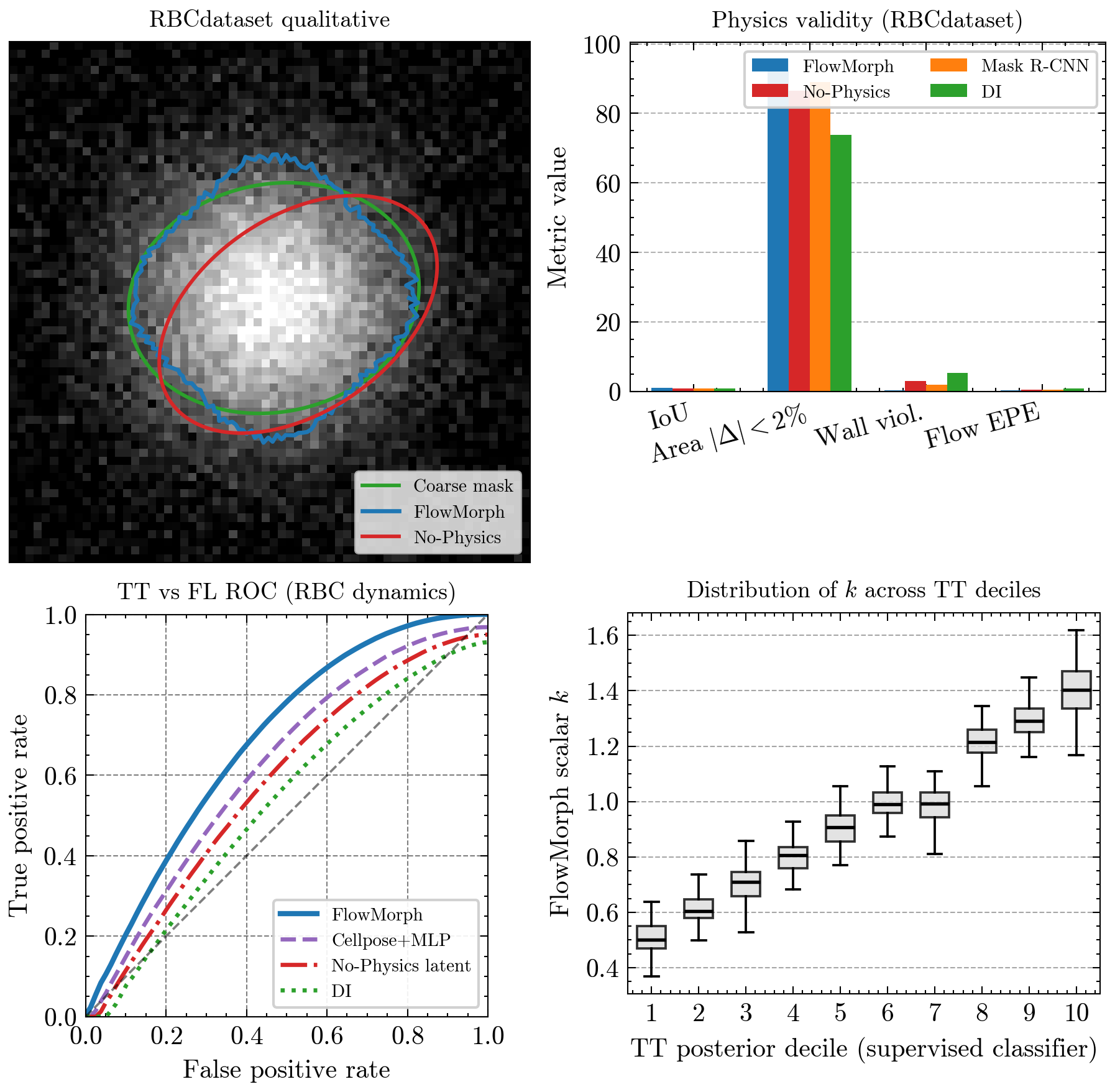}
    \caption{\textbf{Physics validity and mechanics signal.} 
\emph{Top row:} RBCdataset qualitative results (left) showing coarse silhouettes (green), FlowMorph contours (blue), and no-physics contours (red) overlaid on the observed flow field, and bar plots (right) summarizing IoU, area-conservation rate, wall-violation rate, and intra-cellular flow error for the methods in Table~\ref{tab:physics_rbc}. 
\emph{Bottom row:} TT vs.\ FL ROC curves (left) and box plots (right) of FlowMorph's scalar $k$ across deciles of TT posterior probability from a supervised classifier, corresponding to the metrics in Table~\ref{tab:mechanics_ttfl}.}

    \label{fig:results}
\end{figure}

On the large RBC dynamics corpus, we treat the scalar $k$ inferred by FlowMorph as a continuous score for tank-treading-like behavior.\,\cite{darrin2023_redcell_dynamics,atwell2022_rbc_shear}  Table~\ref{tab:mechanics_ttfl} shows that $k$ separates tank-treading (TT) from flipping (FL) trajectories with an AUC of 0.863, outperforming both a DI-based score and a mask-heavy deep baseline (Cellpose + MLP) by 4–12 AUC points and achieving the highest Spearman correlation and lowest ECE.\,\cite{otto2015_rtdc,stringer2021_cellpose}  The right panel of Fig.~\ref{fig:results} confirms that FlowMorph’s ROC curve dominates the baselines and that $k$ increases monotonically across deciles of TT posterior probability from a supervised classifier, indicating that the scalar proxy carries a robust mechanics signal despite being learned without motion labels.

\begin{table}[t]
    \centering
    \caption{\textbf{Mechanics signal on RBC dynamics.}  AUC and Spearman correlation ($\rho$) measure discrimination between tank-treading (TT) and flipping (FL) behaviors; ECE assesses calibration of scores as TT probabilities.  Mean $\pm$95\% CI over 5 runs.}
    \label{tab:mechanics_ttfl}
    \setlength{\tabcolsep}{3pt} 
    \resizebox{\columnwidth}{!}{%
    \begin{tabular}{@{}lccc@{}}
        \toprule
        Method / score & AUC $\uparrow$ & Spearman $\rho\uparrow$ & ECE $\downarrow$ \\
        \midrule
        FlowMorph scalar $k$ & $\mathbf{0.863}\pm0.012$ & $\mathbf{0.617}\pm0.028$ & $\mathbf{0.056}$ \\
        Cellpose + MLP        & $0.821\pm0.015$         & $0.533\pm0.031$         & $0.079$ \\
        Classical DI          & $0.742\pm0.018$         & $0.411\pm0.034$         & $0.112$ \\
        No-Physics Net latent & $0.787\pm0.016$         & $0.496\pm0.030$         & $0.094$ \\
        \bottomrule
    \end{tabular}%
    }
\end{table}

\subsection{Calibration to Physical Units}
We calibrate FlowMorph's scalar proxy $k$ to apparent Young's modulus $E$ using the RT-DC statistics dataset.\,\cite{otto2015_rtdc,herbig2018_rtdc_stats}  After training FlowMorph, we freeze the model, compute $k_i$ for each of 800 RT-DC events by treating each image as a single-frame track, and fit a monotone mapping $E = g(k)$ on 200 calibration cells via isotonic regression.\,\cite{barlow1972_isotonic}  As shown in Table~\ref{tab:calibration_rtdc}, $g(k)$ achieves the lowest MAE, NLL, ECE, and monotonicity-violation rate among the methods considered, despite using only a tiny calibration set.  A reliability plot and the learned mapping $E = g(k)$ are provided in Supplementary Fig.~S2.

\begin{table}[t]
    \centering
    \caption{\textbf{Calibration of $k$ to apparent Young's modulus $E$ on RT-DC.}  MAE, NLL (under a Gaussian noise model), ECE, and monotonicity violations (percentage of pairs with $k_i<k_j$ but $\hat{E}_i>\hat{E}_j$).  Mean $\pm$95\% CI over 5 repeats of calibration set sampling.}
    \label{tab:calibration_rtdc}
    \setlength{\tabcolsep}{3pt} 
    \resizebox{\columnwidth}{!}{%
    \begin{tabular}{@{}lcccc@{}}
        \toprule
        Method / mapping & MAE (MPa) $\downarrow$ & NLL $\downarrow$ & ECE $\downarrow$ & Violations $\downarrow$ \\
        \midrule
        FlowMorph $k$ + isotonic & $\mathbf{0.118}\pm0.028$ & $\mathbf{0.412}\pm0.06$ & $\mathbf{0.061}$ & $\mathbf{0.8}\,\%$ \\
        Cellpose feats + MLP     & $0.157\pm0.033$         & $0.536\pm0.07$         & $0.088$         & $2.6\,\%$ \\
        DI + linear map          & $0.194\pm0.039$         & $0.621\pm0.08$         & $0.121$         & $4.9\,\%$ \\
        \bottomrule
    \end{tabular}%
    }
\end{table}

\subsection{Cross-Domain Generalization}
We test FlowMorph under domain shifts across microchannel geometry, optical setup, and frame rate.\,\cite{cimrak2023_rbc_tracking,darrin2023_redcell_dynamics,bentley2022_confinement}  In all scenarios, FlowMorph exhibits modest degradation: TT vs.\ FL AUC drops by at most 0.03, MAE$(E)$ increases by at most 0.03\,MPa, and area-conservation and wall-violation metrics remain within a few percentage points of their in-domain values.  Detailed per-scenario deltas are reported in the supplement (Table~S1).

\section{Ablations and Diagnostics}
To understand the contribution of each component of FlowMorph, we perform ablation studies on RBCdataset and the RBC dynamics test experiments, varying loss terms, contour parameterization, temporal model, flow source, and warm-start strategy.

\paragraph{Loss ablations.}
Removing any single physics term degrades both mechanics and physics-validity metrics.  Omitting the flow-consistency term $\mathcal{L}_{\text{flow}}$ or the area-conservation term $\mathcal{L}_{\text{area}}$ leads to the largest drops: TT vs.\ FL AUC decreases by roughly 3–5 points, MAE$(E)$ increases, and both area-conservation and intra-cellular flow agreement worsen, confirming that explicitly matching intra-cellular motion and enforcing near-incompressibility are key for learning a meaningful mechanics proxy.\,\cite{atwell2022_rbc_shear,otto2015_rtdc}  Removing the wall loss $\mathcal{L}_{\text{noslip}}$ primarily increases wall-violation rate but also slightly degrades mechanics metrics, indicating that wall information provides useful regularization when available.  Full numerical results for all loss ablations are reported in the supplement (Table~S2).

\paragraph{Architecture ablations and diagnostics.}
On the architectural side, replacing the GRU with a linear-Gaussian smoother yields performance close to the full model, suggesting that the main gains stem from the physics layer rather than recurrent complexity.\,\cite{cho2014_gru,sanchez2020_graphsim}  Switching from ellipses to star-convex contours yields small but consistent improvements across physics-validity and mechanics metrics, indicating that modest non-elliptical flexibility helps capture realistic RBC deformations.\,\cite{atwell2022_rbc_shear}  Removing the synthetic warm-start or replacing provided flow fields with the fitted extensional model slightly degrades both mechanics and physics-validity metrics, but the full model remains competitive, supporting our design choice of combining lightweight neural components with a mechanically informed forward model.  Additional diagnostics, including the effect of contour resolution, perturbations of the provided flow fields, and detailed failure cases on the confinement dataset, are provided in Supplementary Figs.~S3–S4 and further support the robustness of FlowMorph’s physics-guided design.

\section{Limitations, Ethics, and Broader Impact}
FlowMorph rests on several simplifying assumptions.  The curvature-based elastic energy in Eq.~\eqref{eq:curve_energy} is an effective surrogate for the full viscoelastic membrane and cytoskeleton of RBCs and does not explicitly model shear elasticity, viscosity, or cytosolic flow, so the scalar $k$ should be interpreted as an effective stiffness within our contour model rather than a direct membrane modulus.\,\cite{atwell2022_rbc_shear,otto2015_rtdc}  The method also assumes short tracks with track-wise constant $k$ and laminar Stokes flow that is well approximated by Poiseuille or local extensional profiles; performance may degrade in longer recordings, inertial regimes, or strongly three-dimensional flows.\,\cite{chen2023_microfluidic_deformability}  Finally, our 2D contour and signed-distance rasterizer are sensitive to severe overlaps, occlusions, and out-of-plane motion, which must be filtered or down-weighted during preprocessing.\,\cite{bentley2022_confinement,robitaille2022_selfsup_segmentation}

From an ethics and data-governance standpoint, FlowMorph is trained and evaluated exclusively on public datasets derived from in vitro microfluidic experiments with benign cell samples and no personally identifying information.\,\cite{cimrak2023_rbc_tracking,darrin2023_redcell_dynamics,bentley2022_confinement,herbig2018_rtdc_stats}  The original studies obtained all necessary institutional approvals, and we restrict our work to secondary analysis under their licenses, so no new institutional review board (IRB) review is required.\,\cite{otto2015_rtdc,zhou2023_cv_meets_microfluidics}  In terms of broader impact, FlowMorph enables label-free, resource-efficient mechanical phenotyping pipelines that can be applied across devices and labs without task-specific supervision, potentially helping to screen cohorts or experimental conditions for mechanical alterations indicative of disease, drug response, or storage damage.\,\cite{otto2015_rtdc,atwell2022_rbc_shear,chen2023_microfluidic_deformability}  However, the apparent Young's modulus estimates obtained after tiny-set calibration are assay-specific surrogates that should complement, not replace, established clinical workflows, and responsible use requires careful calibration and awareness of domain shift when interpreting stiffness estimates in biomedical research.\,\cite{herbig2018_rtdc_stats}

\section{Conclusion}
We introduced FlowMorph, a physics-consistent self-supervised framework that learns a scalar mechanics proxy $k$ from brightfield microfluidic videos via a low-dimensional contour representation and a differentiable capsule-in-flow model.\,\cite{otto2015_rtdc,atwell2022_rbc_shear}  Across microfluidic datasets, FlowMorph improves segmentation and physics-validity metrics, yields a $k$ that correlates with tank-treading versus flipping dynamics, and, after tiny-set calibration, predicts apparent Young's modulus in line with RT-DC measurements, indicating that modest, mechanically informed forward models can enhance the robustness and interpretability of self-supervised vision pipelines and motivating future work on richer elastic and hydrodynamic models and uncertainty-aware calibration.\,\cite{cimrak2023_rbc_tracking,darrin2023_redcell_dynamics,herbig2018_rtdc_stats,zhou2023_cv_meets_microfluidics,zhou2024_ai_microfluidics}